%% file: main.tex
\documentclass[10pt,twocolumn,letterpaper]{article}

\usepackage[pagenumbers]{cvpr} 
\input{preamble}
\definecolor{cvprblue}{rgb}{0.21,0.49,0.74}
\usepackage[pagebackref,breaklinks,colorlinks,allcolors=cvprblue]{hyperref}
\usepackage[accsupp]{axessibility}
\def\paperID{9046} 
\def\confName{CVPR}
\def\confYear{2026}

\title{FedAFD: Multimodal Federated Learning via Adversarial Fusion and Distillation}

\author{
Min Tan$^{1}$ \quad
Junchao Ma$^{2}$ \quad
Yinfu Feng$^{3}$\thanks{Corresponding authors.} \quad
Jiajun Ding$^{2}$ \quad
Wenwen Pan$^{2}$ \\
Tingting Han$^{2\ast}$ \quad
Qian Zheng$^{4}$ \quad
Zhenzhong Kuang$^{2}$ \quad
Zhou Yu$^{2}$ \\
\small $^{1}$ Zhejiang Key Laboratory of Space
Information Sensing and
Transmission, Hangzhou Dianzi University \\
\small $^{2}$ Laboratory of Complex Systems
Modeling and Simulation, School of Computer
Science and Technology, Hangzhou Dianzi University \\
\small $^{3}$ Alibaba International Digital Commerce Group \\
\small $^{4}$ College of Computer Science and Technology, Zhejiang University  \\
{\tt\small 
\{tanmin,ttinghan\}@hdu.edu.cn, 
fyf200502@gmail.com
}
}

\begin{document}
\maketitle
\input{sec/0_new_abstract}    
\input{sec/1_new_intro}
\input{sec/2_new_relatedworks}
\input{sec/3_new_method}
\input{sec/4_new_experiment}
\input{sec/5_new_conclusion}
\input{sec/6_new_acknowledgement}

{
    \small
    \bibliographystyle{ieeenat_fullname}
    \bibliography{main}
}

\end{document}

%% file: preamble.tex
\usepackage{multirow}
\usepackage{booktabs}
\usepackage{color}
\usepackage{algorithm}
\usepackage{algpseudocode}
\usepackage{newfloat}
\usepackage{listings}
\usepackage{amsmath}
\usepackage{amsfonts}
\usepackage{pifont}
\usepackage{subcaption}
\usepackage{float}

%% file: sec/0_new_abstract.tex
\begin{abstract}
Multimodal Federated Learning (MFL) enables clients with heterogeneous data modalities to collaboratively train models without sharing raw data, offering a privacy-preserving framework that leverages complementary cross-modal information. However, existing methods often overlook personalized client performance and struggle with modality/task discrepancies, as well as model heterogeneity. To address these challenges, we propose FedAFD, a unified MFL framework that enhances client and server learning. On the client side, we introduce a bi-level adversarial alignment strategy to align local and global representations within and across modalities, mitigating modality and task gaps. We further design a granularity-aware fusion module to integrate global knowledge into the personalized features adaptively. On the server side, to handle model heterogeneity, we propose a similarity-guided ensemble distillation mechanism that aggregates client representations on shared public data based on feature similarity and distills the fused knowledge into the global model. Extensive experiments conducted under both IID and non-IID settings demonstrate that FedAFD\footnote{Code: https://github.com/Chao2433/FedAFD} achieves superior performance and efficiency for both the client and the server.
\end{abstract}

%% file: sec/1_new_intro.tex
\section{Introduction}
Multimodal training has garnered interest for its ability to utilize complementary knowledge from various modalities \cite{luo2025graph,wu2025enriching}.
It has enabled significant progress in diverse tasks, including object detection \cite{chen2025comprehensive}, visual question answering \cite{liu2025towards}, and cross-modal retrieval \cite{su2025dica,tan2025disco}. However, due to hardware limitations and high acquisition costs, collecting large-scale multimodal data (\textit{e.g.}, vision-language pairs) for centralized model training is often impractical. While collaborative training across multiple data holders provides a viable alternative to build powerful models, direct data sharing exacerbates privacy risks, especially when fusing sensitive cross-modal correlations from distributed sources. This dilemma has sparked a significant demand for privacy-preserving multimodal federated learning (MFL), where decentralized participants jointly train models without exposing their raw data. In particular, MFL offers a trustworthy collab for training large-scale multimodal foundation models across domains\cite{tan2019image,tan2023electromagnetic}, especially under the increasing emphasis on data privacy in the era of large models\cite{wang2025cluster}. 
\begin{figure}[t]
\centering
\includegraphics[width=\linewidth]{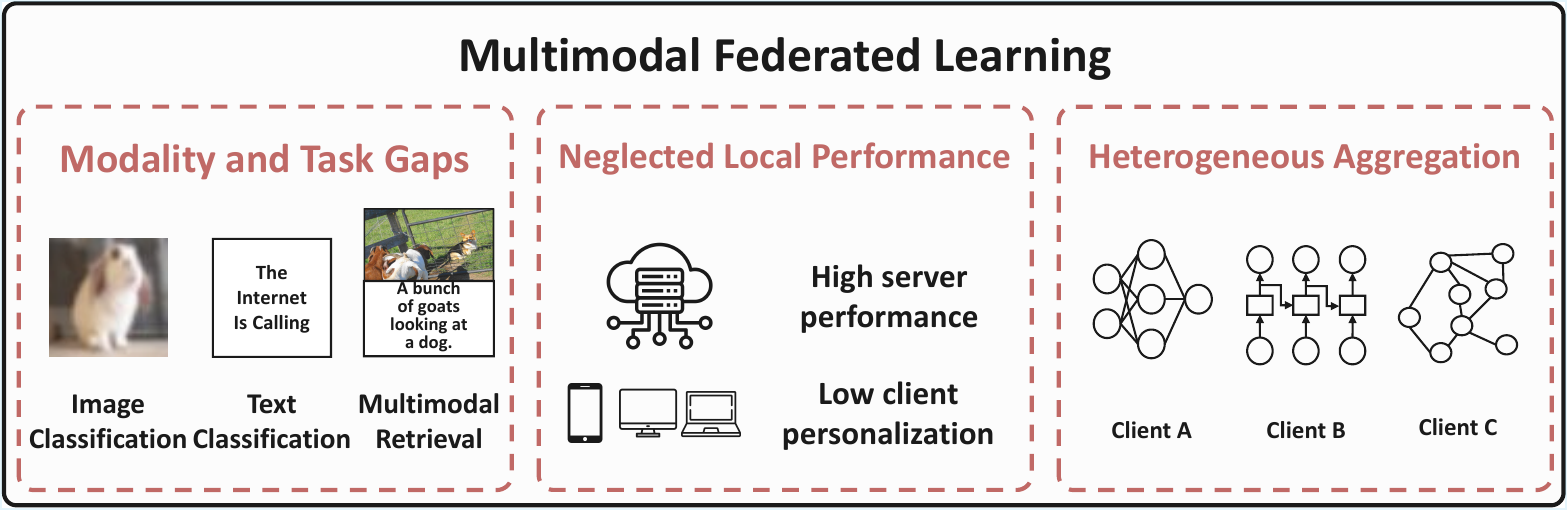}
\caption{The challenges of multimodal federated learning}
\label{intro}
\end{figure}

Despite recent progress in federated learning (FL) \cite{mcmahan2017communication}, most existing works focus on unimodal clients \cite{yan2025fedvck,zheng2025confree,yi2025pfedes} and the existing MFL approaches \cite{xiong2022unified,chen2024breaking} often assume homogeneous clients with identical modalities and tasks, an assumption that rarely holds in real-world scenarios where participants naturally exhibit modality and task heterogeneity. Moreover, model heterogeneity arising from diverse client architectures exacerbates these challenges (Fig.~\ref{intro}), further complicating collaborative training. Some methods~\cite{yu2023multimodal} attempt to address these challenges and improve global model performance,, trading off local performance in the process. These limitations hinder the practical application of MFL in real-world scenarios, thereby underscoring the urgent need for an effective edge-cloud collaboration framework that simultaneously enhances both global and local model performance.

These critical problem motivates our central research question: How can we design an MFL framework that harmonizes diverse modalities and tasks, synergistically enhances both global and local models, and facilitates effective collaboration across heterogeneous participants?

To address these problems, we present a novel multimodal federated learning framework named FedAFD, which systematically tackles the above three challenges through a three-stage design. Specifically, we introduce a modality-task bi-level adversarial alignment module that employs two complementary adversarial losses to align feature distribution across both modality and task levels, effectively mitigating model drift. Building on the aligned cross-domain features, we further design a granularity-aware feature fusion mechanism that adaptively integrates global semantics and personalized cues at multiple feature levels, thereby enhancing local representation quality and improving personalization through the implicit incorporation of commonsense. To transfer client-side knowledge to the server under architectural heterogeneity, we propose a similarity-guided ensemble distillation strategy that adaptively aggregates local representations based on the semantic consistency with the global representations and distills the fused knowledge into the global model. 

Our main contributions are three-fold: i) We propose a novel federated learning framework, FedAFD, that fully leverages complementary information across tasks and modalities to simultaneously enhance models of both diverse edge devices and the cloud server; ii) FedAFD is the first framework to jointly address cross-modal/task alignment, task-aware personalization, and architecture-agnostic aggregation in a unified manner;
and iii) Extensive experiments demonstrate that FedAFD outperforms SOTA methods under both IID and Non-IID settings.

\begin{figure*}[t]
\centering
\includegraphics[width=1\textwidth]{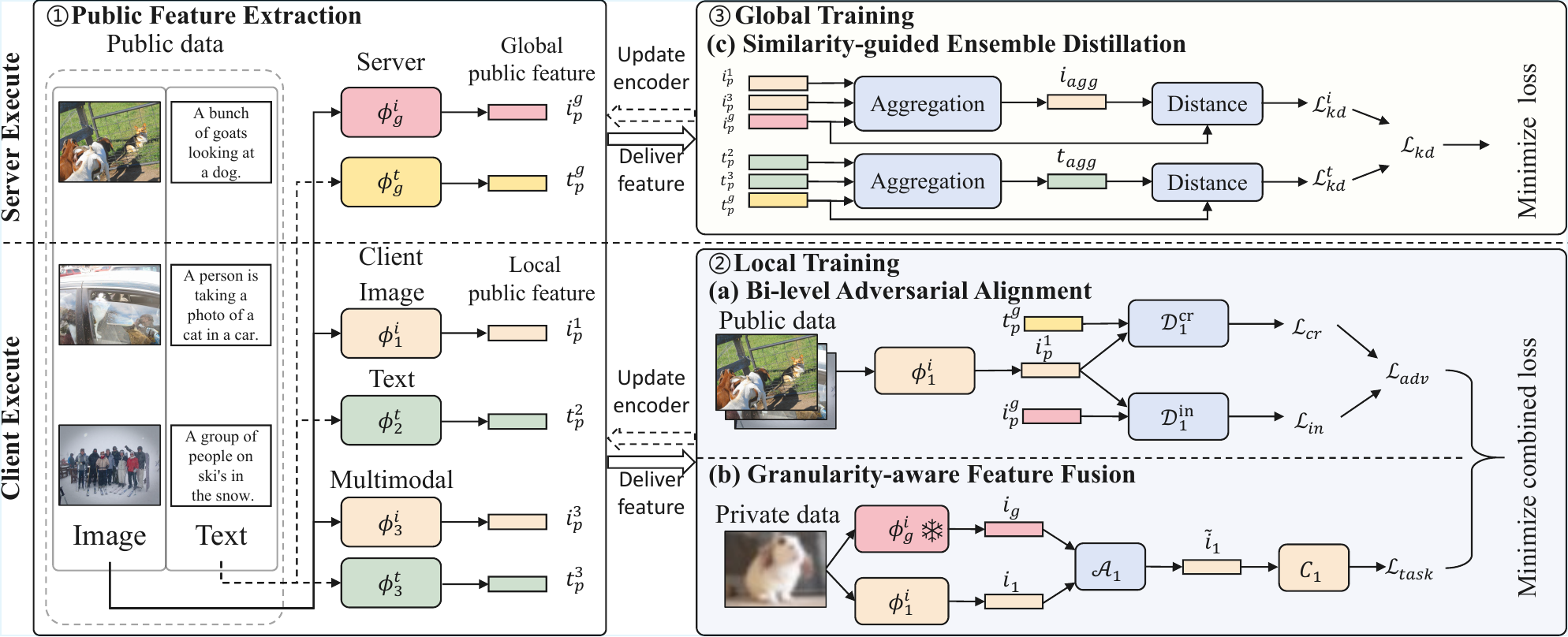}
\caption{Overview of our proposed FedAFD comprising three steps: \ding{172} Server is trained on a public dataset and extracts global public features.
\ding{173} With delivered global representations and encoders, clients train local models on private data by granularity-aware local and global fusion, enhanced with bi-level adversarial feature alignment.
\ding{174} Clients extract local public features of the public dataset and then sends it to the server. The server performs adaptive aggregation and updates the global model via similarity-guided ensemble distillation.
}
\label{framework}
\end{figure*}
\label{sec:intro}

%% file: sec/2_new_relatedworks.tex
\section{Related Works}
\subsection{Multimodal Federated Learning}
Multimodal federated learning enables different clients with different modalities to collaboratively optimize models through local training without sharing raw data. Early MFL studies assume a homogeneous setting, where all clients share identical tasks and model architectures. MMFed~\cite{xiong2022unified} employs a unified model structure across clients and fuses different modalities through attention mechanisms. FedSea~\cite{tan2023fedsea} introduces selective feature alignment to handle Non-IID multimodal data. Recent works have begun addressing modality heterogeneity. CreamFL \cite{yu2023multimodal} employs intra-modal and inter-modal contrastive regularization terms to mitigate modality and task gaps. Another line of research focuses on handling missing or incomplete modalities across clients. FedMBridge~\cite{chen2024fedmbridge} proposes a bridgeable representation space to connect heterogeneous multimodal clients. PEPSY~\cite{nguyen2025learning} learn reconfigurable representations to support multimodal federated learning under missing data scenarios. FedMosaic~\cite{seo2025not} propose a personalized federated learning framework tailored to heterogeneous multimodal clients. Despite these advances, the model drift caused by modality and task heterogeneity in multimodal federated learning remains underexplored.

\subsection{Feature Fusion}
Feature fusion is a widely adopted technique in traditional machine learning and deep learning, aiming to integrate features from diverse sources or modalities to improve model performance. It has shown effectiveness in a variety of tasks, including image deraining \cite{chen2023hybrid}, visual recognition \cite{li2024feature}, image retrieval \cite{wu2023asymmetric}, object detection \cite{huang2025l4dr}, multimodal tasks \cite{zhao2023cddfuse}, and change detection \cite{deng2022feature,deng2023tchange}. Recently, several works have explored feature fusion in the context of federated learning \cite{tasbaz2024feature,feng2023fedmultimodal} by incorporating the complementary information to boost model performance. Still, they do not consider the data heterogeneity. Although some FL methods attempt to address data imbalance \cite{yue2023specificity,cao2024federated}, they have not explicitly tackled the feature fusion problem under modality and task gaps in multimodal federated learning. Furthermore, they fail to achieve a balance between local personalization and global generalization, which is crucial for effective collaboration in real-world applications.

\subsection{Knowledge Distillation in Federated Learning}
Knowledge distillation~\cite{hinton2015distilling} has been widely adopted in federated learning to mitigate model heterogeneity and performance degradation under non-IID data. 
Early studies primarily achieve consensus at the logit level using public data. FedMD~\cite{li2019fedmd} aligns the outputs of heterogeneous local models via consensus distillation. 
Subsequent works enhance aggregation through more sophisticated weighting strategies, including variance-aware ensembling in FedET~\cite{cho2022heterogeneous} and entropy-based weighting in FedGEMS~\cite{cheng2021fedgems}. More recent advances extend distillation to representation-level and sample-aware paradigms. CreamFL~\cite{yu2023multimodal} aggregates cross-modal representations through contrastive learning. FedMKD~\cite{li2024resource} employs attention mechanisms to weight multi-teacher representations. FedDFA~\cite{wang2025fed} introduces boundary-aware weighting by assigning larger distillation weights to samples near decision boundaries. 
Despite these advances, most existing approaches remain confined to unimodal settings and seldom investigate explicit interactions between local and global representations or dynamic representation-level integration.

%% file: sec/3_new_method.tex
\section{The Proposed Model: FedAFD}
We illustrate our overall framework in Fig.~\ref{framework}, where the global and local models are iteratively trained. In each communication round, clients first receive the server's representation of the public data and update their local models using bi-level adversarial alignment and granularity-aware feature fusion. Then, clients generate the public data representations based on their own modality and transmit them to the server. Finally, the server performs similarity-guided dynamically weighted representation aggregation and ensemble knowledge distillation to update the global model. The whole training procedure is illustrated in Algorithm \ref{algorithm1}. We provide the training framework in Appendix A.1.

\subsection{Preliminary}
We consider a multimodal federated learning setting involving three types of clients: $N_I$ unimodal image clients for image classification, $N_T$ unimodal text clients for text classification, and $N_M$ multimodal clients for image-text retrieval. Specifically, each image client $c \in \{1, \ldots, N_I\}$ holds a private dataset $\mathcal{I}_c = {(x_c^k, y_c^k)}_{k=1}^{|\mathcal{I}_c|}$, where $x_c^k$ is the $k$-th image and $y_c^k$ is its corresponding label. Each text client $c \in \{N_I+1, \ldots, N_I+N_T\}$ maintains a private dataset $\mathcal{T}_c = {(x_c^k, y_c^k)}_{k=1}^{|\mathcal{T}_c|}$, consisting of text inputs and their associated labels. Multimodal clients $c \in \{N_I+N_T+1, \ldots, N_I+N_T+N_M\}$ are equipped with private datasets $\mathcal{M}_c = {(x_c^k, y_c^k)}_{k=1}^{|\mathcal{M}_c|}$, where each sample is a semantically aligned image-text pair. In addition, a public multimodal dataset $\mathcal{P} = {(x_p^k, y_p^k)}_{k=1}^{|\mathcal{P}|}$, also composed of image-text pairs, is available to both the server and all clients. Each client trains a personalized model $f_c(\cdot;w_c)$ on its private dataset. For unimodal clients, $w_c=\{\phi_c (\cdot), \mathcal{C}_c (\cdot)\}$, where $\phi_c(\cdot)$ is the modality-specific encoder, $C_c(\cdot)$ is the local classifier. For multimodal clients, $w_c=\{\phi^i_c(\cdot),\phi^t_c(\cdot)\}$, where $\phi^i_c(\cdot)$ and $\phi^t_c(\cdot)$ are the image and text encoders, respectively. All clients collaboratively train a server model $f_g(\cdot;w_g)$ for image-text retrieval task, parameterized by $w_g=\{\phi^i_g(\cdot),\phi^t_g(\cdot)\}$, where $\phi^i_g(\cdot)$ and $\phi^t_g(\cdot)$ are the global image and text encoders, respectively.


\subsection{Model Structure}
To tackle modality and task gaps, limited local personalization, and heterogeneous model aggregation in multimodal federated learning, we propose three key modules shown in Fig.~\ref{framework}: 1) Bi-level adversarial alignment aligns local and global representation distribution within and across modalities through adversarial training; 2) Granularity-aware feature fusion harmonizes the capabilities of local and global models to enhance personalization; and 3) Similarity-guided ensemble distillation aggregates knowledge from local models and updates the global model via modality-aware knowledge distillation. Collectively, these modules enable robust and adaptive federated learning.
\begin{algorithm}[t]
\caption{The Learning of FedAFD} 
\label{algorithm1}
\textbf{Input}: {Public dataset $\mathcal{P}$, local dataset $\mathcal{D}_c$ of the $c$-th client, communication rounds $T$, local epochs $E$}, server model $f_g(\cdot;w_g)$, client model $f_c(\cdot;w_c)$ which contain feature extractor $\phi_c$, intra-modal discriminator $\mathcal{D}_c^{in}$, cross-modal discriminator $\mathcal{D}_c^{cr}$, GFF module $\mathcal{A}_c$, and local classification $\mathcal{C}_c$.\\
\textbf{Output}: {The final server model $w_g^{T}$
\begin{algorithmic}[1]
\State \textbf{Server executes:}
\For{$t = 0$ \textbf{to} $T-1$}
    \State $w_g^{t} \gets w_g^t - \eta \nabla \mathcal{L}(\mathcal{P}; w_g^t)$
    \State $(i_p^g,t_p^g) \gets f(\mathcal{P};w_g^t)$
    \State Send global features $i_p^g,t_p^g$ and encoder $\phi_g$ to clients
    \State $(i_p^c, i_p^c, w_c^{t+1}) \gets$ \textbf{ClientUpdate}$(c, t, i_p^g, t_p^g, \phi_g)$
    \State $w_g^{t+1} \gets w_g^t-\eta\gamma\nabla\mathcal{L}_{kd}(i_p^g, t_p^g,i_p^c,t_p^c)$
\EndFor
\Function{\textbf{\textup{ClientUpdate}}}{$c, t, i_p^g, t_p^g, \phi_g$}
    \For{$i = 0$ to $E-1$}
        \State $(\mathcal{D}_c^{in},\mathcal{D}_c^{cr})^{i+1} \gets (\mathcal{D}_c^{in},\mathcal{D}_c^{cr})^i+\eta_c\beta\nabla\mathcal{L}_{adv}$
        \State $(\mathcal{A}_c,\mathcal{C}_c)^{i+1} \gets (\mathcal{A}_c,\mathcal{C}_c)^i-\eta_c\nabla\mathcal{L}_{task}$
        \State $\phi_c^{i+1} \gets \phi_c^i-\eta_c(\nabla\mathcal{L}_{task}+\beta\nabla\mathcal{L}_{adv})$
    \EndFor
    \State $(i_p^c, t_p^c) \gets f_c(\mathcal{P};w_c^{t,E})$
    \State \Return $i_p^c, t_p^c, w_c^{t,E}$
\EndFunction
\end{algorithmic}}
\end{algorithm}
\subsubsection{Bi-level Adversarial Alignment: BAA}



Multimodal FL faces a fundamental challenge of representation inconsistency due to modality and task heterogeneity. For example, unimodal image clients are trained solely for image classification without exposure to textual data, while multimodal clients are trained on image-text retrieval. They might encode the same images into different latent spaces. This inconsistency can lead to model drift, hindering effective knowledge aggregation and collaborative learning. We treat it as a federated domain adaptation problem: the feature space learned by each client is considered the source domain, while the server's representation space is the target domain. Inspired by domain adaptation theory~\cite{ben2010theory}, the performance of a model on the target domain is bounded by the distribution difference between the source and target domains, this leads us to use adversarial learning to directly minimize the difference in representation distribution between the client and server. The theoretical analysis is provided in Appendix B.

Taking the unimodal image client $c$ as an example(the process is analogous for other clients), in each round of communication, $c$ first receives the global image features $i^g_p$ and global text features $t^g_p$($p$ represents the public data, $g$ represents the global model) generated by the global model for public data. For the $k$-th public data, the local encoder $\phi_c(\cdot)$ generates the image representation $i_p^{c,k}$ , which needs to be aligned with the two types of global representations. As shown in Fig.\ref{framework}(a), to achieve this bi-level alignment, each client is equipped with two adversarial discriminators: a cross-modal discriminator $\mathcal{D}_c^{cr}$ that differentiates between local and global representations from a different modality (\textit{e.g.}, $i^{c,k}_p$ and $t^{g,k}_p$), and an intra-modal discriminator $\mathcal{D}_c^{in}$ that distinguishes between local and global representations of the same modality (\textit{e.g.}, $i^{c,k}_p$ and $i^{g,k}_p$). The specific architecture of the discriminator is provided in Appendix A.2.1. It takes a feature vector as input and outputs a scalar probability indicating the likelihood that the feature originates from the server's global distribution rather than the local client. $\mathcal{D}_c^{in}$,$\mathcal{D}_c^{cr}$ and local encoder are trained in an adversarial manner. Specifically, we first train $\mathcal{D}_c^{in}$ and $\mathcal{D}_c^{cr}$ to identify whether the feature comes from the client or the server. We then train the encoder part of the client to confuse them. This is achieved through a bi-level adversarial loss comprising intra-modal and cross-modal objectives:
\begin{equation}
\left\{ \begin{array}{l}
\mathcal{L}_{\text{in}}^{k}=\log \mathcal{D}_c^{\text{in}}(i^{g,k}_p) +\log(1-\mathcal{D}_c^{\text{in}}(i^{c,k}_p)),\\
\mathcal{L}_{\text{cr}}^{k}=\log \mathcal{D}_c^{\text{cr}}(t^{g,k}_p) +\log(1-\mathcal{D}_c^{\text{cr}}(i^{c,k}_p)),\\
\end{array} \right.
\end{equation}
where $i^{g,k}_p$ = $\phi_g^i(x_p^k)$, $t^{g,k}_p$ = $\phi_g^t(x_p^k)$ and $i^{c,k}_p$ = $\phi_c(x_p^k)$. The eventual adversarial loss is defined as:
\begin{equation}
\mathcal{L}_{adv}=\frac{1}{|\mathcal{P}|}\sum_{k=1}^{|\mathcal{P}|}(\mathcal{L}_{\text{in}}^{k}+\mathcal{L}_{\text{cr}}^{k}).
\label{eq:ADVLOSS}
\end{equation}
It optimizes adversarially, with discriminators maximizing and encoders minimizing Eq.~(\ref{eq:ADVLOSS}). This bi-level adversarial mechanism reduces the representation distribution difference within and across modalities between clients and the server, effectively mitigating model drift and guiding local encoders to generate personalized representations enriched with shared commonsense knowledge. 

\subsubsection{Granularity-aware Feature Fusion: GFF}
Local features contain knowledge relevant to the local task, whereas global features encapsulate generalized semantics. While the BAA module aligns feature distributions between clients and the server, it may introduce excessive global knowledge, leading to a decrease in local performance. To dynamically balance this specialization-generalization trade-off, we introduce the Granularity-aware Feature Fusion (GFF) module at the sample level.

 


Inspired by~\cite{dai2021attentional}, GFF employs an attention-based method to fuse complementary local and global features adaptively. Taking a unimodal image client $c$ as an example (the process is analogous for other clients), for each local epoch, client $c$ performs local training on its own private dataset $\mathcal{I}_c$. As shown in Fig.\ref{framework}(b), the $k$-th input image $x^k_c$ is encoded by both the local image encoder $\phi_c$ and the global image encoder $\phi^i_g$ provided by the server, producing the local feature $i_c^k$ and $i_g^k$ ($c$ represents the local encoder and $g$ represents the global encoder, respectively). Feature fusion is carried out in two steps. First, an attention-based gating mechanism adaptively integrates the local and global features to produce an intermediate fused representation:
\begin{equation}
    h_c^k = M(i_c^k + i_g^k) \otimes i_c^k + (1 - M(i_c^k + i_g^k)) \otimes i_g^k.
\end{equation}
where $\otimes$ denotes element-wise multiplication, and $h_c^k$ is the intermediate fused feature. Next, we refine this output through another attention stage to generate the final fused feature $\widetilde{i}_c$, the feature fusion module $\mathcal{A}_c$ is defined as:
\begin{equation}
    \widetilde{i}_c^k = \mathcal{A}_c(i_c^k,i_g^k)=M(h_c^k) \otimes i_c^k + (1 - M(h_c^k) \otimes i_g^k.
\end{equation}
The attentional weights $M(\cdot)$ is defined as:
\begin{equation}
    M(x)=\sigma(T_1(x)+T_2(x)),
\end{equation}
where $x$ is the input feature vector, $\sigma$ denotes the sigmoid activation, and $T_1(\cdot)$ and $T_2(\cdot)$ are two parallel nonlinear transformations to capture multi-scale contextual information defined as in \cite{dai2021attentional}, the implementation is provided in Appendix A.2.2. 
The final fused feature $\widetilde{i}_c$ is used to compute the client's task-specific loss:
\begin{equation}
    \mathcal{L}_{task}=\frac{1}{|\mathcal{I}_c|}\sum_{k=1}^{|\mathcal{I}_c|}l(\widetilde{i}_c^k,y_c^k;w_c),
\label{eq:losstask}
\end{equation}
where $l(\widetilde{i}_c^k,y_c^k;w_c)$ represents the task-specific objective function (\textit{e.g.}, cross-entropy loss). Eq.(\ref{eq:losstask}) encourages each client to adaptively integrate personalized local and generalized global knowledge, thereby boosting client-specific model performance.

\subsubsection{Similarity-guided Ensemble Distillation: SED}
To effectively aggregate knowledge from heterogeneous local models into the global model, we propose a Similarity-guided Ensemble Distillation (SED) module that dynamically assigns aggregation weights to local representations based on the semantic consistency between the local and global representations. We consider a representation to be more reliable if it is closer to the global representation of the same public data in feature space, while being distinguishable from unrelated data.

As shown in Fig.\ref{framework}(c), after local training, each client generates feature representations on the public dataset and uploads them to the server. The server then acts as the student, while the participating clients collectively serve as teachers. Taking the image modality as an example, for the $k$-th public data, let $i_p^{c,k}$ denote the image representation generated by client $c$’s encoder, and $i_p^{g,k}$ denote the corresponding image representation produced by the server’s global encoder. The similarity score is computed as:
\begin{equation}
s^{c,k} = \log \frac{\exp(sim(i_p^{c,k},i_p^{g,k}))}{\sum_{j=1}^{|\mathcal{P}|} \exp(sim(i_p^{c,k},i_p^{g,j}))},
\end{equation}
where $|\mathcal{P}|$ denotes the number of public datas, $sim(\cdot)$ represents the cosine similarity between two representations. The normalized aggregation weights across all clients with image modality for sample $k$ are then computed as:
\begin{equation}
w^{c,k} = \frac{\exp(s^{c,k})}{\sum_{c' \in \pi_{\text{img}}} \exp(s^{c',k})},
\end{equation}
where $\pi_{\text{img}}$ indexes the clients equipped with the image modality. The corresponding aggregated teacher representation for sample $k$ is then computed as:
\begin{equation}
i_{agg}^k = \sum_{c \in \pi_{\text{img}}} w^{c,k} \cdot i_p^{c,k}.
\end{equation}
The server distills knowledge from the clients by minimizing the L2-distance between the aggregated teacher and the global student representation:
\begin{equation}
\mathcal{L}_{kd}=\frac{1}{|\mathcal{P}|}\sum_{k=1}^{|\mathcal{P}|}\left({||i^k_{agg}-i^{g,k}_{p}||}_2+{||t^k_{agg}-t^{g,k}_{p}||}_2\right).
\end{equation}
This representation-level distillation process enables the server to extract knowledge from heterogeneous client models without requiring parameter-level consistency. By leveraging similarity-based ensemble aggregation, SED effectively addresses the challenge of model heterogeneity and enhances the server’s representation capacity.

%% file: sec/4_new_experiment.tex
\section{Experiments}
In this section, we first describe the experimental setup and baselines, then present performance comparisons between FedAFD and other methods, along with ablation study and interpretability analysis. Due to the space limitation, further hyperparameter analysis and communication cost analysis are provided in Appendix C.

\subsection{Experimental Setup}
\begin{table*}
\centering
\begin{tabular*}{\hsize}{@{}@{\extracolsep{\fill}}lccc cc c c@{}}
\toprule
\multirow{3}{*}
{Size}&\multicolumn{4}{c}{Clients}&\multicolumn{3}{c}{Server}
\\\cline{2-5} \cline{6-8}&CIFAR-100&AGNEWS&\multicolumn{2}{c}{Flickr30k}&\multicolumn{3}{c}{MS-COCO}\\
\cline{2-2} \cline{3-3} \cline{4-5} \cline{6-8}
&acc@1&acc@1&i2t R@1&t2i R@1&i2t R@1&t2i R@1&rsum R@1\\
\hline
10k&\textbf{33.18}&\textbf{51.98}&32.48&25.68&33.98&26.18&60.16\\
20k&\underline{32.03}&\underline{50.81}&\underline{34.50}&\underline{27.39}&\underline{39.90}&\underline{31.91}&\underline{71.81}\\
30k&30.08&50.35&\textbf{36.88}&\textbf{28.64}&\textbf{43.66}&\textbf{34.43}&\textbf{78.09}\\
\bottomrule
\end{tabular*}
\caption{FedAFD with varying public data sizes. The evaluation metric is top-1 accuracy(acc@1) for CIFAR-100/AGNEWS classification, top-1 Recall(R@1) for Flickr30k and MS-COCO cross-modal retrieval (i2t/t2i).}
\label{public_size}
\end{table*}

\begin{table*}[t]
\centering
\begin{subtable}[t]{\textwidth}
\centering
\begin{tabular*}{\hsize}{@{}@{\extracolsep{\fill}}lccc cc c cc@{}}
\toprule
\multirow{3}{*}{Method}&\multicolumn{4}{c}{Clients}&\multicolumn{4}{c}{Server}
\\\cline{2-5} \cline{6-9}&CIFAR-100&AGNEWS&\multicolumn{2}{c}{Flickr30k}&\multicolumn{4}{c}{MS-COCO}\\
\cline{2-2} \cline{3-3} \cline{4-5} \cline{6-9}
&acc@1&acc@1&i2t R@1&t2i R@1&i2t R@1&t2i R@1&rsum R@1&\#rounds\\
\hline
LOCAL&\underline{46.52}&85.56&\underline{30.70}&23.95&32.62&25.01&57.63&29\\
\midrule
FedMD~\cite{li2019fedmd}&37.36&85.54&25.90&21.45&32.74&25.52&58.26&25\\
FedGEMS~\cite{cheng2021fedgems}&37.33&85.63&25.98&21.02&33.04&25.64&58.68&26\\
FedET~\cite{cho2022heterogeneous}&46.44&\underline{86.07}&30.65&\underline{24.21}&32.92&25.88&58.80&23\\
CreamFL~\cite{yu2023multimodal}&34.01&85.22&26.45&20.72&33.08&\underline{25.98}&\underline{59.06}&\underline{22}\\
FedMKD~\cite{li2024resource}&41.01&84.63&30.63&23.98&33.34&25.71&59.05&\underline{22}\\
FedDFA~\cite{wang2025fed}&32.28&84.83&27.70&22.73&\underline{33.36}&25.48&58.84&25\\
\midrule
FedAFD&\bf{61.04}&\bf{89.34}&\bf{36.55}&\bf{29.83}&\bf{33.78}&\bf{26.02}&\bf{59.80}&\textbf{21}\\
\bottomrule
\end{tabular*}
\caption{Results under IID data setting.}
\label{IID}
\end{subtable}
\begin{subtable}[t]{\textwidth}
\centering
\begin{tabular*}{\hsize}{@{}@{\extracolsep{\fill}}lccc cc c cc@{}}
\toprule
\multirow{3}{*}{Method}&\multicolumn{4}{c}{Clients}&\multicolumn{4}{c}{Server}
\\\cline{2-5} \cline{6-9}&CIFAR-100&AGNEWS&\multicolumn{2}{c}{Flickr30k}&\multicolumn{4}{c}{MS-COCO}\\
\cline{2-2} \cline{3-3} \cline{4-5} \cline{6-9}
&acc@1&acc@1&i2t R@1&t2i R@1&i2t R@1&t2i R@1&rsum R@1&\#rounds\\
\hline
LOCAL&28.07&48.35&22.33&\underline{18.44}&32.48&25.06&57.54&29\\
\midrule
FedMD~\cite{li2019fedmd}&22.54&48.18&19.13&15.63&33.00&25.47&58.47&25\\
FedGEMS~\cite{cheng2021fedgems}&22.84&48.30&18.93&16.05&33.12&25.50&58.62&27\\
FedET~\cite{cho2022heterogeneous}&\underline{31.86}&\underline{49.38}&\underline{22.63}&18.22&33.20&25.72&58.92&27\\
CreamFL~\cite{yu2023multimodal}&22.14&42.16&18.38&15.49&\underline{33.72}&25.89&\underline{59.61}&\underline{21}\\
FedMKD~\cite{li2024resource}&24.99&47.99&22.33&18.37&33.24&\underline{25.94}&59.18&\underline{21}\\
FedDFA~\cite{wang2025fed}&23.09&43.79&19.68&17.13&33.56&25.54&59.10&26\\
\midrule
FedAFD&\textbf{33.18}&\textbf{51.98}&\textbf{32.48}&\textbf{25.68}&\textbf{33.98}&\textbf{26.18}&\textbf{60.16}&\textbf{20}\\
\bottomrule
\end{tabular*}
\caption{Results under Non-IID data setting.}
\label{NONIID}
\end{subtable}
\caption{Performance comparison with baselines on diverse clients and the server under both IID and Non-IID settings. \#rounds denotes the communication round number when they are close to convergence on the server, \textit{i.e.}, rsum R@1 equals 57.50.}
\label{overall}
\end{table*}

Following the experimental setting in~\cite{yu2023multimodal}, we simulate a realistic multimodal federated learning scenario where heterogeneous clients with either unimodal (\textit{e.g.}, vision or text) or multimodal data collaborate through a central server while maintaining strict data privacy. The goal is to apply federated learning across these heterogeneous clients to collaboratively train a larger global model on the server, aiming to optimize global tasks.

\subsubsection{Datasets}
We construct local clients for image classification (CIFAR-100), text classification (AGNEWS), and multimodal retrieval (Flickr30k), while the global server focuses on cross-modal retrieval using MS-COCO. The datasets and their configurations are described below.

\textbf{CIFAR-100} \cite{krizhevsky2009learning}: CIFAR-100 contains 60K 32×32 color images across 100 classes. 
Under IID setting, each client samples uniformly across all classes; While for the Non-IID setting, we use a Dirichlet distribution ($\alpha$=0.1) to simulate label distribution skew.

\textbf{AGNEWS} \cite{zhang2015character}: It consists of 120K training and 7.6K test samples across four classes, 
with IID/non-IID partitioning identical to CIFAR-100.

\textbf{Flickr30k} \cite{plummer2015flickr30k}: Flickr30K comprises approximately 31K images, each annotated with five captions, and is split into 29K for training, 1K for validation, and 1K for testing. Under the IID setting, training samples are evenly distributed, while the Non-IID setting uses shard-based splitting based on distinct image groups.

\textbf{MS-COCO} \cite{chen2015microsoft}: MS-COCO consists of 123,287 images, with 113,287, 5000, and 5000 images for training, validation, and testing, respectively. Each is paired with five annotated captions. To facilitate efficient communication, we randomly sample a subset of the data to construct a public dataset shared between the server and all clients.



\subsubsection{Evaluation Metrics}
We evaluate performance using task-specific metrics: Top-1 accuracy(acc@1) for CIFAR-100/AGNEWS classification, Top-1 Recall(R@1) for Flickr30k image-to-text (i2t) and text-to-image (t2i) cross-modal retrieval task. Server performance averages five 1K-test-image folds, reporting Top-1 Recall(R@1) for MS-COCO and their sum(rsum). Besides performance, we test efficiency by measuring the number of communication rounds needed for the server to reach a target result. The best and second-best results are in bold and underlined, respectively.



\subsubsection{Baselines}
We compare FedAFD with several state-of-the-art FL approaches including:FedMD~\cite{li2019fedmd}, FedGEMS~\cite{cheng2021fedgems}, FedET~\cite{cho2022heterogeneous}, CreamFL~\cite{yu2023multimodal}, FedMKD~\cite{li2024resource}, FedDFA~\cite{wang2025fed}. LOCAL denotes the setting where each client and the server are trained independently without collaboration. We extend all uni-modal baselines to multimodal federated learning scenarios for a fair comparison. Specifically, for FedMD, we introduce a global model on the server; each client performs local knowledge distillation using representations of public data generated by the server model. The client representations are then aggregated on the server in proportion to the number of samples each client holds. For FedGEMS, we follow a similar distillation framework but weight the aggregation of local representations by the entropy of their predictions. For FedET, the server performs average model aggregation and aggregates representations according to their variance. For FedMKD, we obtain the aggregated representation through an attention mechanism, which serves as a positive sample for contrastive distillation. For FedDFA, we calculate the entropy of representation and assign a greater distillation weight to samples closer to the decision boundary.

\subsubsection{Implementation Details}
We construct three unimodal image clients, three unimodal text clients, and four multimodal clients. Following standard practice \cite{yu2023multimodal}, we use ResNet-18 \cite{he2016deep} and GRU \cite{chung2014empirical} as backbone models for unimodal image and text clients, respectively. For multimodal clients and the server, we implement the PCME framework \cite{chun2021probabilistic} to support cross-modal retrieval. Specifically, ResNet-18 and ResNet-101 are used as the image encoders for multimodal clients and server, while GRU and BERT-base \cite{devlin2019bert} are used as the corresponding text encoders. The adversarial coefficient $\beta$ and the distillation coefficient $\gamma$ are set to 0.5 and 0.4, respectively. Both image and text features are projected to 256 dimensions.

For optimization, unimodal image and text clients use stochastic gradient descent (SGD) with a batch size of 256 and an exponentially decaying learning rate scheduler. The initial learning rates are set to 5e-2 for image clients and 2e-2 for text clients. Multimodal clients and the server are optimized using the AdamP optimizer with cosine learning rate scheduling, a batch size of 128, and initial learning rates of 1e-5 and 2e-4, respectively.

The training process involves 40 communication rounds between clients and the server, each consisting of 5 local training epochs, resulting in a total of 200 local update steps. To ensure reproducibility and fair comparison across all experiments, we maintain identical random seed initialization for all experiments.


\begin{table*}[t]
\centering
\begin{tabular*}{\hsize}{@{}@{\extracolsep{\fill}}lccc cc c c@{}}
\toprule
\multirow{3}{*}{Method}&\multicolumn{4}{c}{Clients}&\multicolumn{3}{c}{Server}
\\\cline{2-5} \cline{6-8}&CIFAR-100&AGNEWS&\multicolumn{2}{c}{Flickr30k}&\multicolumn{3}{c}{MS-COCO}\\
\cline{2-2} \cline{3-3} \cline{4-5} \cline{6-8}
&acc@1&acc@1&i2t R@1&t2i R@1&i2t R@1&t2i R@1&rsum R@1\\
\hline
FedAFD&\underline{33.18}&\bf{51.98}&\bf{32.48}&\bf{25.68}&\bf{33.98}&\bf{26.18}&\bf{60.16}\\
\midrule
w/o BAA&\bf{33.56}&49.03&\underline{32.13}&25.56&33.70&25.59&59.29\\
w/o GFF&24.94&44.46&22.23&18.31&\underline{33.88}&\underline{25.84}&\underline{59.72}\\
w/o SED&32.21&\underline{50.20}&31.38&\underline{25.59}&33.82&25.74&59.56\\
\bottomrule
\end{tabular*}
\caption{Ablation Study of FedAFD under Non-IID Setting}
\label{Ablation}
\end{table*}

\subsection{Overall Performance}
Before comparative analysis, we examine the influence of public data scale on FedAFD's performance by evaluating the framework with 10K, 20K and 30K public samples under Non-IID settings. Experimental results in Table~\ref{public_size} demonstrate that expanded public datasets enable the server to learn more precise modality-aligned representations through additional image-text pairs, particularly enhancing performance for multimodal clients handling similar tasks. However, unimodal clients may experience marginal performance degradation as they align with fewer task-relevant global features. We utilize 10,000 public data points in the following experiments to reduce memory overhead for all compared approaches.

The overall comparison of proposed FedAFD with several baselines is reported in Table~\ref{overall}, both client- and server-side tasks under both the IID and Non-IID settings are evaluated.
The table shows that: 1) While many existing federated learning methods demonstrate strong server-side performance, their client-side accuracy often falls below even local training (``LOCAL'') baselines. This phenomenon reveals a key limitation: conventional global optimization approaches frequently sacrifice local personalization capability when improving server models. In contrast, FedAFD maintains competitive server-side performance while consistently enhancing client-side accuracy, achieving superior balance between global coordination and local adaptation; 2) Compared to IID settings, FedAFD shows significantly greater improvements over baselines in Non-IID scenarios, highlighting the robustness and demonstrating its substantial potential for real-world heterogeneous data; and 3) In terms of efficiency, FedAFD requires the fewest communication rounds to reach target levels, indicating that its adaptive client-server collaboration mechanism effectively mitigates client drift and accelerates convergence.


\subsection{Ablation Study}
We conduct ablation studies by individually removing BAA (denoted as ``w/o BAA''), GFF (denoted as ``w/o GFF''), and SED (denoted as ``w/o SED'') modules from the complete FedAFD model under Non-IID setting. The comparison results are reported in Table~\ref{Ablation}. It can be observed that: 1) BAA effectively mitigates modality/task gaps, improving server performance while slightly compromising client personalization due to tighter alignment with global representations. Note that removing the BAA module may yield a slight and reasonable performance gain (+0.38\% on CIFAR-100), since BAA enforces local-to-global feature distribution alignment, and excessive global context may marginally weaken local discriminative ability; 2) Removing GFF causes significant performance drops mainly on the edge side, as clients lose the ability to fuse global features, underscoring GFF’s role in enhancing local semantics and boosting server performance through high-quality aggregation; and 3) Replacing SED with simple averaging degrades performance, confirming that SED’s importance-weighted distillation optimally integrates local knowledge for superior global model performance.

\begin{figure}[t]
\centering
    \includegraphics[width=\linewidth]{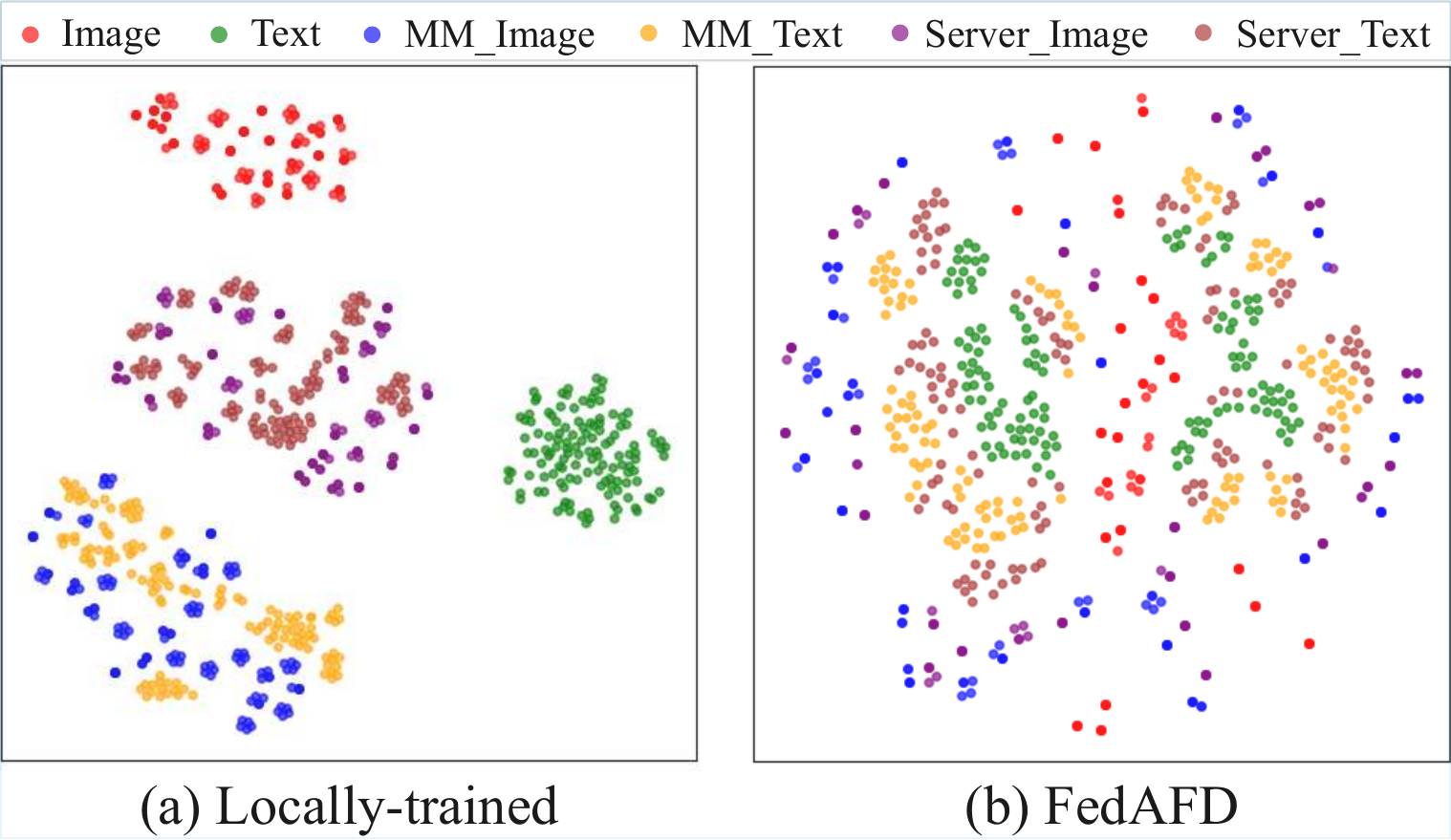}
\caption{T-SNE analysis of feature discrepancy on public data. Locally-trained and FedAFD encoders are compared.}
\label{ours}
\end{figure}

\begin{figure}[t]
\centering
\includegraphics[width=\linewidth]{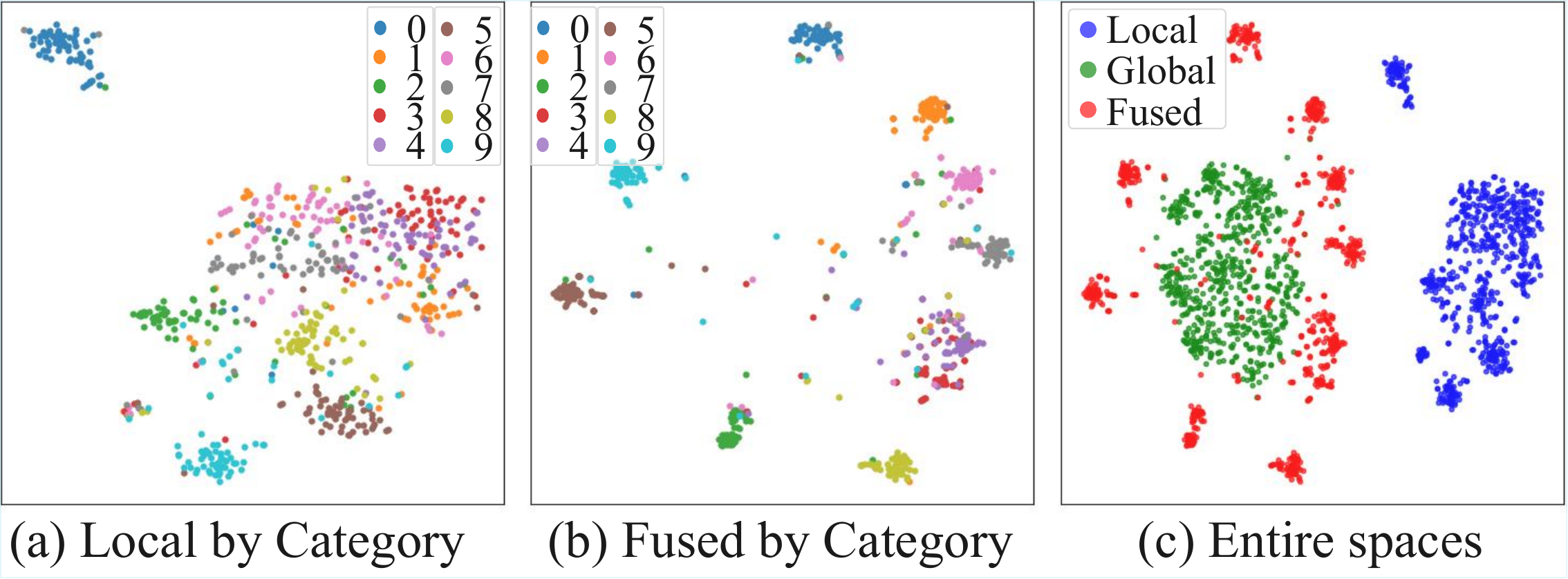}
\caption{T-SNE visualization of different features on CIFAR-100, in terms of both image category and feature types.}
\label{GFF}
\end{figure}



\subsection{Interpretability Analysis}
We provide feature-level interpretability to illustrate our improvements on both the client and server sides.

Firstly, we randomly select 128 samples from the public dataset and compare their representations extracted by models trained under two settings: standalone local training and our proposed FedAFD framework. These samples are passed through the encoders of an image client, a text client, a multimodal client, and the global server model. As shown in Fig.~\ref{ours}(a), in the Locally-trained setting, the extracted features of the same samples are widely scattered across different clients and the server, revealing straightforward task and modality gaps among participants. In contrast, after applying FedAFD, the features from all clients and the server become more aligned and form a compact cluster in the embedding space. This confirms that FedAFD effectively bridges the modality and task discrepancies, achieving semantic consistency for shared samples and enabling better global model adaptation across heterogeneous clients.

Moreover, to validate the impact of the GFF module on client-side features, we conduct a comparative t-SNE visualization analysis using the CIFAR-100 dataset. As shown in Fig.~\ref{GFF}, we visualize features from the first 10 classes under three settings: using the local encoder only (``Local''), using the global encoder (``Global''), and using the GFF-based fusion (``Fused''). Fig.~\ref{GFF}(a) and Fig.~\ref{GFF}(b) show class-wise distributions, while Fig.~\ref{GFF}(c) presents an overall comparison.
The results reveal that the fused features exhibit more precise inter-class boundaries than ``Local'', demonstrating that GFF effectively improves local feature discriminability by adaptively integrating personalized local and generalized global semantics. Notably, the fused features remain close to the local ones, indicating that GFF enhances local performance without compromising personalization. This confirms that FedAFD balances global knowledge integration with client-specific task optimization.

%% file: sec/5_new_conclusion.tex
\section{CONCLUSION}
This paper presents FedAFD, a novel multimodal federated learning framework that addresses three core challenges: modality and task discrepancies, limited local personalization, and server performance degradation under model heterogeneity. We propose bi-level adversarial alignment for unified representation learning, granularity-aware feature fusion to strengthen local encoding, and similarity-guided ensemble distillation for effective knowledge transfer to the global model. Extensive experiments demonstrate that FedAFD consistently improves both client and server performance, offering a scalable and effective solution for real-world multimodal federated learning.


%% file: sec/6_new_acknowledgement.tex
\section{Acknowledgement}
This work was supported by National Natural Science Foundation of China (No.62472133, No.62422204, No.62206082, No.62372147, No.62402152), Zhejiang Province Natural Science Foundation of China (No.LMS26F020018, No.LQN25F020017, No.LQK26F020005, No.LRG26F020001), Key Research and Development Program of Zhejiang Province No.2025C01026 and Scientific Research Innovation Capability Support Project for Young Faculty.